\documentclass[11pt]{article}

\usepackage{palatino,setspace,graphicx,rotating,mathrsfs,bm,listings,caption}
\usepackage{subfigure,latexsym,amsmath,amssymb,natbib, rotating,
  ocg-p}
\usepackage[pdftex,colorlinks=true,citecolor=blue,raiselinks=false]{hyperref}


\begin{document}

\begin{center}
{\LARGE {\bf The Hidden Frame}} \\ {\large How Large Language Models Impact Democratic Society}

\vspace*{5mm}

Wendy K. Tam\footnote{Wendy K. Tam is Professor and Stevenson Chair in the Departments of Political Science, Computer Science, Biomedical Informatics, and the Law School at Vanderbilt University, and an affiliate of the National Center for Supercomputing Applications at the University of Illinois at Urbana-Champaign.  Email: {\sl wendy.k.tam@vanderbilt.edu}}
\end{center}

\begin{abstract}
\begin{singlespace}
Large language models are rapidly becoming an interface between citizens and political information.  They are often regarded as ``a better Google.'' While this analogy might work for some instances, it is unintuitively problematic for democratic politics.  A search engine retrieves human-authored documents, while a language model generates novel text that necessarily embeds invisible framing decisions.  Because conveying knowledge involves framing, a system that generates answers cannot serve as a neutral conduit to ``all human knowledge.''  Instead, these systems are becoming a new kind of political intermediary.  Mechanistic evidence shows that partisan identity is encoded as a locatable geometric direction inside the Llama 3.1 8B model, and that alignment training masks rather than removes this structure.  Building on that evidence, we use the model's training cutoff in 2023 to make those frames visible.  This cutpoint auspiciously falls just before a dramatic realignment in American politics marked by the second Trump administration and the MAHA transformation of health politics, providing us with a natural experiment.  We find that the model presents temporally contingent partisan alignments as {\em knowledge}, with no reliable mechanism for distinguishing fact from opinion.  Because citizens judge political claims by who made them and when, an answer that carries neither cue hampers the judgment on which self-governance depends.  Large language models purporting to summarize ``all human knowledge'' are, in actuality, simply magnifying the cultural and partisan divides inherent in their training data.
\end{singlespace} \end{abstract}

\pagestyle{headings}

\clearpage
\newpage

\section{Introduction}

Every transformative communication technology has remade the polity that adopted it, often with unintended consequences.  The printing press broke a clerical monopoly on textual authority and helped ignite a century of religious wars before it underwrote the Enlightenment~\citep{Eisenstein:79}.  The telegraph and broadcast media each reorganized who could speak to whom, and on what terms~\citep{Starr:04}.  Cable television dissolved the shared national audience of the broadcast era and allowed citizens to sort themselves by interests~\citep{Prior:07}.  Social media created echo chambers by further fueling ideological sorting with algorithmic amplification of engagement-maximizing content~\citep{Sunstein:17, Pariser:11}.  Repeatedly, in history, we have witnessed how advances in communication infrastructure expanded beyond the technical arena to reallocate power by defining what a society knows~\citep{ProcacciniTam:27}.

Large language models (LLMs) are the newest such tectonic shift in communication infrastructure.  They are often regarded as ``a better Google'' even though the two technologies fundamentally differ.  A search engine {\em retrieves} human-created content whereas a language model {\em generates} text that no human wrote, through internal computations that remain, at best, only partially understood.  To be sure, LLMs are not simply an incremental improvement in information retrieval.  They are, at once, the new curators and the new distributors of information, as well as something no prior technology has ever been.  They are its producers.

Language models have indisputable benefits and enormous reach.  They can be patient tutors, help plan meals, write code, review legal contracts, and explain medical diagnoses.  The technology is transformative and the effects are palpable across many aspects of daily life.  Recently, citizens have begun to consult chatbots about elections, asking them what is on the ballot and whom to vote for~\citep{Medina:26, Truong:26}.  Our focus here is not on whether language models write code or tutor math competently, but on what happens to democratic society when these models answer the questions through which a public forms its understanding of its politics.

LLMs are the newest {\em intermediary} sitting between society and what it knows.  Like the wire service, newspapers, and broadcast media before them, they fill the role of transforming an unstructured world of information into digestible snippets.  Their functioning as an intermediary is easy to overlook because language models do not present themselves as such.  Trained on much of what humanity has committed to writing, they present themselves as repositories of ``all human knowledge,'' ready to convey that knowledge to whoever asks.  Behind that claim, however, stands no newsroom, no editor, and no masthead.  Society is quietly trading its institutional intermediaries for apps.

But knowledge does not exist in a platonic, presentation-free form that a sufficiently advanced machine could simply retrieve.  Answering questions about politics differs from answering a question about a calculus problem.  Although math problems may be approached in multiple ways, these approaches nevertheless converge on a single verifiable answer.  In contrast, a question about politics has no single answer that is stored in some training corpus waiting to be retrieved.  Answering political questions requires compressing far more information than any generated answer could contain, as well as deciding what information to convey and how to convey it.  That selection is precisely what information intermediaries have done and what language models now do.

Communicating anything {\em requires} selection and emphasis.  These are value-laden choices that shape how information is perceived.  A system that answers questions therefore cannot be a neutral window into ``all human knowledge.''  To be sure, conveying knowledge necessarily involves framing.  But, what determines that frame?

For a newspaper, framing choices lie with editors and owners.  For a network, producers and sponsors fill this role.  These frames are visible.  Readers know if they are holding the {\em New York Times} or the {\em New York Post}, and viewers know if they are watching Fox News or MSNBC.  A byline tells readers who researched and wrote a story.  A dateline tells them whether the story is current or outdated.  A language model's frame, however, does not originate in editorial choice but in model training.  It forms as the model absorbs its training corpus, which encompasses the writing of many people over many years.  Developers then align the model with ``human values.''   No bylines or datelines survive to tell a reader whose frame it is or when it was formed.

The question---how intermediaries select and frame the world---has been studied for over a century, from~\citet{Lippmann:22}'s pseudo-environment through agenda setting and framing~\citep{IyengarKinder:87, Entman:93} to the political economy of post-broadcast media~\citep{Prior:07} and the measurement of online segregation~\citep{GentzkowShapiro:11, Guess:21}.  However, the tools developed to answer it assume human-authored content whose sources could be coded.  But unlike their predecessors, language models do not merely select and summarize human-authored content.  Instead, they generate their own answers, leaving no trace of an outlet or author whose leanings can be inferred.  Because each answer is generated anew, there is also no fixed text that an audience shares or that a researcher could sample.  Moreover, reading the outputs alone cannot reveal where their frames originate.

Open weight models restore some of this purchase by allowing us to open and experimentally manipulate the model.  Earlier work showed whose frames a model carries~\citep{Tam:26a}.  A model's training cutoff lets us see when they were fixed.  Llama 3.1 was released in July 2024, with a pretraining corpus that ends in December 2023~\citep{Dubeyetal:24}, just before a tumultuous period of political realignment.  With Donald Trump's return to the presidency in January 2025, much of what the model had encoded as settled policy was rapidly overturned.  A frame is hardest to see when the world still agrees with it, but this realignment auspiciously affords a rare vantage point for understanding how model's frame information.

We begin with the argument that conveying knowledge requires framing.  Generated output from language models strips away the cues that allow its readers to recognize framing.  We then describe the partisan geometry of the model and use the training cutoff as a natural experiment to date its frame, before turning to how the composition of the training corpus shapes that frame. Finally, we consider the impact on democratic society when its citizens rely on intermediaries that frame information without an author or a date.

\section{Conveying Knowledge Necessarily Involves Framing}
\label{sec:framing}

Frameless communication simply does not exist.  This is among the oldest lessons of the study of public opinion.  \citet{Lippmann:22} observed that citizens respond not to the world but to the pictures in their heads, pictures assembled by intermediaries who must select and simplify because ``the real environment is altogether too big, too complex, and too fleeting for direct acquaintance.''  \citet{Mannheim:36} generalized the point, holding that knowledge is socially situated.  \citet{Schattschneider:60} gave the same observation its political edge when he wrote that ``[t]he definition of the alternatives is the supreme instrument of power.''  In short, whoever decides which considerations are relevant to a question has already shaped its answer.  Indeed, \citet{Heidegger:54} located framing in technology itself, describing it as a mode of revealing that brings some aspects of the world into view while holding others back.

More recent literature turned these insights into measurable claims.  To frame, in \citet{Entman:93}'s canonical definition, is ``to select some aspects of a perceived reality and make them more salient in a communicating text, in such a way as to promote a particular problem definition, causal interpretation, moral evaluation, and/or treatment recommendation\dots''  Decades of experimental work have shown that these selections move opinion.  Emphasizing free speech rather than public safety changes tolerance for a hate rally~\citep{NelsonClawsonOxley:97}.  Emphasizing costs rather than coverage changes support for health reform.  Small changes in how an issue is presented can produce large changes in opinion~\citep{ChongDruckman:07}.  Importantly, framing effects do not require false statements.  Every fact in a framed account can be true.  The politics is in the selection and the emphasis.  Nor is framing a defect of communication that better journalism, or that better engineering, could eliminate.  Rather, it is a necessary component of communicating anything at all.  A summary must choose what to put first.  An answer must choose which considerations count, which authorities are worth citing, which uncertainties deserve mention, and what to leave out. Knowledge can exist without these choices.  Conveying knowledge cannot.

The aspiration to frameless communication is nonetheless persistent, and its history is instructive.  Twentieth-century American journalism institutionalized objectivity as its professional creed, but the working definition that emerged was procedural rather than metaphysical.  Reporters were trained to attribute claims as well as to report the opposing side.  Objectivity was a {\em method} for managing frames, not a way to elude them.  Its practitioners understood as much.  The ``view from nowhere'' is a regulative ideal, not an attainable position~\citep{Nagel:86}.  Even the genres we generally regard as neutral are nevertheless framing devices.  For instance, an encyclopedia entry must decide what belongs in the first sentence, which controversies merit an entry, and how to organize the headings.  The closer a text comes to being regarded as ``reference,'' the more its framing choices seem to disappear.  That disappearance is an illusion, but a consequential one, since it is what makes the reference register politically potent.

Importantly, the distinction between fact and opinion is likewise not a property of sentences.  Nothing in the syntax of ``the program would cost \$32 trillion'' distinguishes it from ``the program would be a disaster.''  That difference is maintained by social machinery.  The separation of news pages from editorial pages, peer review, corrections columns, and professional sanctions for fabrication help sort claims to epistemic categories and police the boundaries between them.  When that machinery is absent, the distinction loses its enforcement mechanism.  Indeed, human readers lean on this machinery constantly.  We consider, for instance, whether a claim appeared in a wire report or an op-ed, and whether it appeared this year or five years ago.  We have become so accustomed to these cues that we barely notice our reliance on them.


What varies across information regimes is not {\em whether} framing occurs but {\em where} it occurs.  In the print and broadcast eras, framing was performed by identifiable human institutions, and a reader knew, or could learn, that an editorial in one newspaper leaned one way and an editorial in another leaned another way.   The search engine complicated this structure without dissolving it.  While its rankings are themselves a form of framing, since deciding which results appear first shapes what a reader sees and potentially weighs most heavily~\citep{EpsteinRobertson:15}, the documents it returnes still carried bylines, mastheads, genres, and dates, so a reader could tell whose framing each reflects and when it was written.  Even with this embedded ranking, framing remained distributed across a plural ecosystem of visible authors.

LLM generation, however, collapses the oldest apparatus we have for judging what to believe.  For as long as claims have circulated, we have weighted them by who made them, and how and when they were made.  When a language model answers a question, it makes every framing decision identified by the literature inside its own computations.  It selects the considerations and chooses the authorities.  It sets the emphasis and calibrates the confidence.  The result arrives as one authorless text.  While sources may be cited, the user nevertheless cannot ask {\em who} wrote the generated output, for {\em which} outlet, {\em when}, or with {\em what} orientation.  Generated output has no byline.  Nor does it carry a date, though it is pinned to one all the same since every corpus has a closing date.  The frames a model learns from its training data do not extrapolate further.  The frames have not disappeared per se.  They still exist, but they have moved inside the model, where they are invisible.  Nothing in the output marks whose frame it is or from when.  Exploring the answers to these questions requires looking under the hood of the model.
 
\section{The Partisan Geometry of a Language Model}
\label{sec:geo}

As a transformer language model reads a prompt, it breaks the text into tokens, which are words or word fragments, and converts each token into a long list of numbers, known as an activation vector.  It revises those numbers through each layer of computation.  In Llama 3.1 8B, each vector contains 4,096 numbers and passes through 32 layers.  At any stage of processing, the text can be pictured as a set of points in a 4,096-dimensional space.  These vectors are the model's internal representation of the text, the substrate from which it generates its response, one token at a time.  In an open weight model like Llama 3.1 8B, these internal vectors are fully observable.

Recent work mines these activation vectors to explore whether the model's internal state encodes a notion of partisanship.  Using 190,491 tweets from sitting members of Congress, labeled by party,~\citet{Tam:26a} trained a logistic regression probe to predict an author's party from the model's internal activations alone.  By Layer 18, about halfway through the network, that probe separates Republican from Democratic text with an AUC of 0.945 and a Cohen's $d$ of 1.94, a gap of nearly two standard deviations.  The probe's weights define a single direction in this 4,096-dimensional space, perpendicular to the boundary that best separates Republican from Democratic text.  Projecting an activation onto that direction places it along a line running from Democratic to Republican.  Partisanship, in other words, can be read off a single direction in the activation space.  This direction is the {\em partisan axis}.

\begin{figure}[htbp]
  \centering
  \includegraphics[width=\textwidth]{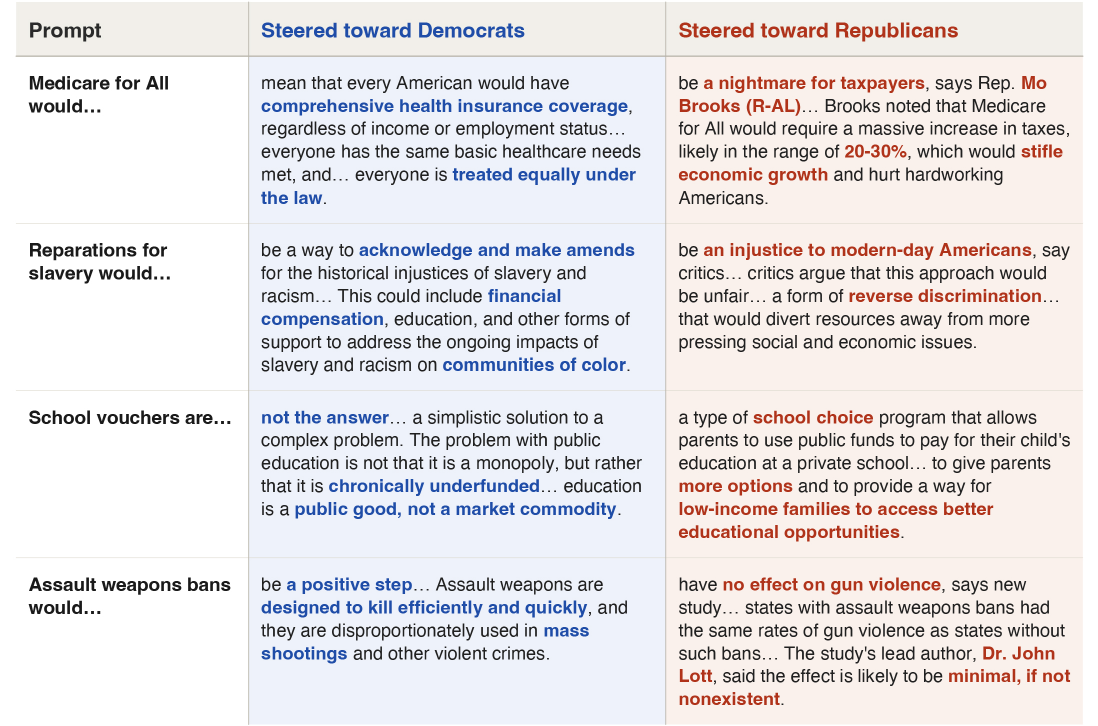}
  \caption{Steering along the partisan axis. Identical prompts, completed by
  the same model, under steering toward the Democrats (left) and toward the
  Republicans (right).}
  \label{fig:steer}
\end{figure}

Finding such an axis tells us only that the model's internal representations encode some notion of partisanship.  That is, there are activation values that the model registers as Republican or Democratic.  It does not tell us whether or how the model uses that representation.  The axis might be something the model merely tracks rather than a lever that it pulls.  To see what the axis does, we submit a prompt to the model.  We let the computation proceed as normal until it reaches Layer 18.  At that point, we shift the activation values toward the positive, or Republican, end of the partisan axis.  The model then continues its computation as usual and produces an answer.  Next, we enter the identical prompt again, this time shifting the activations toward the negative, or Democratic, end.  The model produces a different answer.  Comparing the two answers shows how the output changes when the activations move along the partisan axis.  This intervention is called steering.

Figure~\ref{fig:steer} shows the effect of steering on the response to prompts about four contested issues.  Steered toward the Republican pole, ``Medicare for All would\dots'' completes as ``a nightmare for taxpayers.''  Steered the other way, the same prompt yields ``comprehensive health insurance coverage regardless of income or employment status.''  The prompt is identical and the model is the same.  Only the position along the partisan axis has changed, so the difference follows directly from moving a single internal lever.  The axis does not merely correlate with partisanship. It causally drives the partisanship of what the model says.

The partisan axis was learned from a fixed training corpus, {\em at a particular moment in time}.  If it is a snapshot, it should encode which positions and authorities belonged to which party when that text was collected, but nothing outside the training corpus.  The model's training cutoff offers a way to test this proposition.

\section{The Training Cutoff Provides a Natural Experiment}
\label{sec:frozen}

The partisan geometry in the Llama 3.1 model was developed from a training corpus that ended in December 2023.  How consequential that cutoff is for answering queries depends on how far the politics the model encoded has since moved.  For most of the postwar era, partisan change was incremental, operating on the scale of decades.  Critical elections punctuate long periods of stability~\citep{Key:55}, and issue evolutions, in which the parties exchange positions on an issue or from consensus to cleavage, unfold across multiple electoral cycles as elites reposition and mass publics slowly follow~\citep{CarminesStimson:89}.  Thus, a language model whose picture of politics lags reality by two years would ordinarily miss little of structural importance.

The period since 2024, however, is unique and especially rich in political upheaval.  Between the close of the model's training corpus and the time of this writing, as a result of Donald Trump returning to the presidency in January 2025, American politics has undergone one of its most rapid realignments in living memory.  Among his administration's first acts was the dismantling of federal diversity, equity, and inclusion programs by executive order~\citep{EO14151:25}, converting what had been an institutionalized bureaucratic consensus into a terminated program and a partisan battleground.  The TikTok divestiture law, enacted in April 2024 (after the pretraining cutoff), briefly shut the platform down in January 2025 before the new administration suspended enforcement~\citep{PAFACA:24}.  The partisan positions surrounding the ban scrambled, with Trump, who had first sought to ban the platform in 2020, recasting himself as its protector.  In March 2025, the administration paused military aid to Ukraine~\citep{Reuters:25ukraine}, marking a Republican reversal on what had been, through the corpus period, a broadly bipartisan commitment.

Health politics were also dramatically upended.  Robert F. Kennedy Jr., whose vaccine skepticism had placed him outside the elite mainstream of both parties for two decades, was confirmed as Secretary of Health and Human Services in February 2025.  Under the Make America Healthy Again (MAHA) banner, positions that the pre-2024 corpus identified as fringe became the official policy of a Republican administration.  The CDC's Advisory Committee on Immunization Practices was dismissed wholesale and reconstituted~\citep{CNN:26acip}.  The number of diseases targeted by vaccines recommended for every child was cut from seventeen to eleven~\citep{NBC:26schedule}.  In September 2025, over the objection of the American College of Obstetricians and Gynecologists (ACOG), the administration announced that the FDA would warn physicians of a claimed link between prenatal acetaminophen use and autism~\citep{HHS:25tylenol}.  Indeed, health politics did not merely drift, its partisan valence partially {\em inverted}.  Skepticism of pharmaceutical products and of institutional medicine, historically distributed across both parties' fringes, became mainstream right-coded governing policy, while defense of the medical institutions, such as ACOG and the American Academy of Pediatrics (AAP), became opposition-coded.

For our purposes, then, the realignment that followed the model's training corpus cutoff provides the basis for a research design.  The training cutoff functions as a sharp temporal cutpoint.  Everything the model encodes about these topics derives from a world in which RFK Jr.\ was a bipartisan pariah, DEI was federal policy, Ukraine aid was a bipartisan commitment, and prenatal Tylenol use was a fringe worry.  By steering the model along its partisan axis on prompts about these topics, we can observe how a frozen partisan geometry behaves when the political world it compressed has dramatically changed.

 The post-cutoff shifts manifested in three main forms, with each predicting a different signature in a frozen model.  The first is {\em valence inversion}.  The issue persists and the contending authorities persist, but the assignment of positions to the partisan camps, as in the case of health policy, flips.  The second is {\em consensus collapse}, where a position held broadly across the elite spectrum during the corpus period becomes contested or abandoned by one camp.  Examples include aid to Ukraine, the pre-2024 rejection of Kennedy's vaccine views, and the institutional consensus around federal DEI programs.  The third is {\em scrambling}, where events reshuffle coalitions in ways that map onto no prior axis.  TikTok provides an example where a ban was first sought by one Republican administration, enacted under a Democratic one, and then suspended when Trump returned.  A model frozen before these shifts occurred should exhibit a distinct failure mode for each, namely inverted steering on the first, same-side steering on the second, and uniform adoption of the old common framing on the third.  Under consensus collapse, the converged answer lands on what is now one camp's position, whereas under scrambling it lands on a framing that maps onto neither camp today.

\section{The Temporal Fingerprint Dates the Frame}
\label{sec:fingerprint}

\begin{figure}[htbp]
  \centering
 \includegraphics[width=\textwidth]{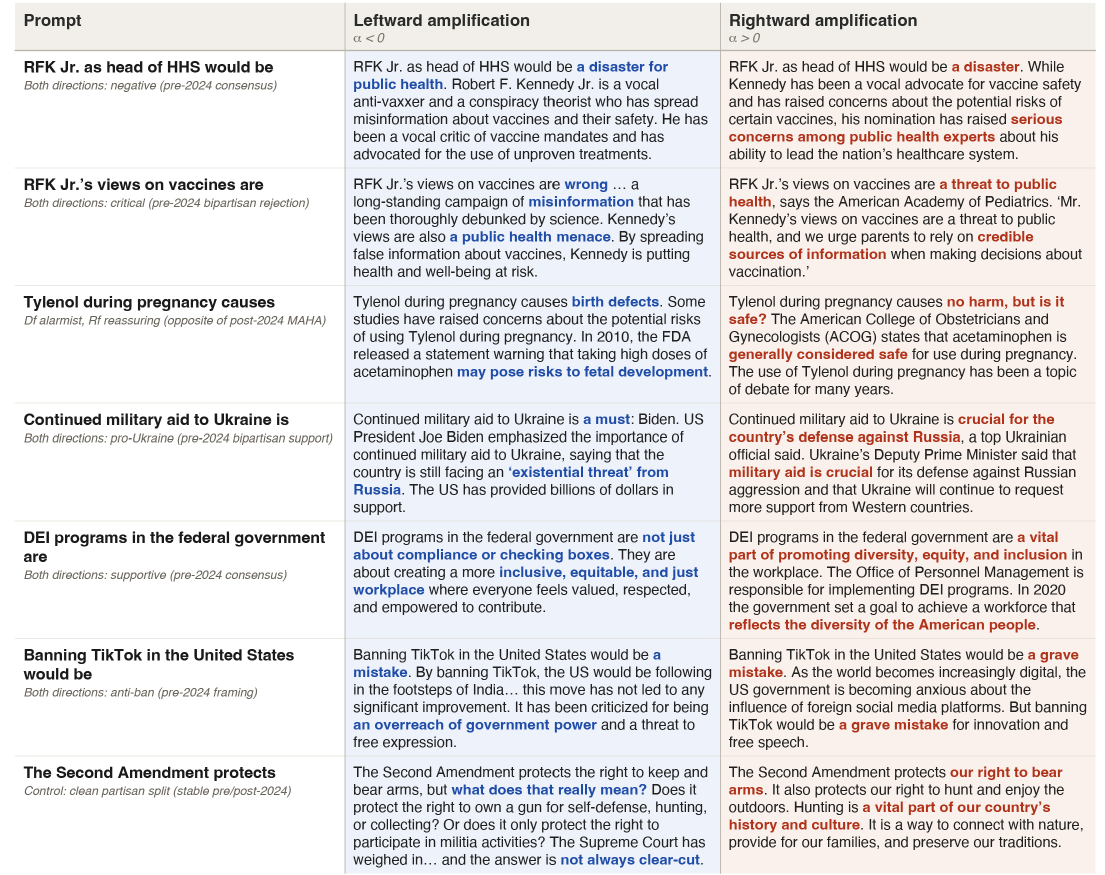}
 \caption{The temporal fingerprint.  Bidirectional amplification along the partisan axis for six prompts touching post-cutoff realignments, plus one stable-cleavage control (Second Amendment).  On realigned topics, completions reflect pre-2024 alignments.  Both steering directions land on the same side where a consensus collapsed or coalitions scrambled, and the assignment runs backwards when valence is inverted.}
 \label{fig:temporal}
\end{figure}

Steering along the partisan axis reveals how the model applies partisan frames to its answers.  On topics that realigned after the cutoff, it also reveals when those frames were fixed. Figure~\ref{fig:temporal} presents the results for six realigned topics and one stable control.  The pattern is striking.  On every topic where an elite consensus collapsed or coalitions scrambled, steering toward {\em either} partisan pole produces output from the {\em same} side, the side that prevailed in the training window.  Steered leftward on ``RFK Jr.\ as head of HHS would be\dots,'' the model completes the prompt with ``a disaster for public health,'' describing Kennedy as ``a vocal anti-vaxxer and a conspiracy theorist.''  Steered {\em rightward}, the model also completes ``a disaster,'' noting that the nomination ``has raised serious concerns among public health experts.''  The rightward intervention, the same intervention that elsewhere conjures Republican consultants and Cato Institute studies, cannot produce a defense of the sitting Republican administration's health secretary, because no such defense exists in the model's corpus.  The pre-2024 elite consensus, in which Kennedy was a bipartisan pariah, is the only material the geometry contains.  The same holds for his views on vaccines.  Leftward steering calls them ``a public health menace.''  Rightward steering attributes ``a threat to public health'' to the American Academy of Pediatrics.

The Tylenol prompt is revealing because the realignment inverted the valence rather than merely shifting it.  Steered leftward on ``Tylenol during pregnancy causes\dots,'' the model produces the alarmist completion, warning of ``birth defects'' and citing FDA statements about fetal development.  Steered rightward, it produces the reassuring completion, declaring acetaminophen ``generally considered safe'' on the authority of ACOG.  This mapping is exactly backwards relative to the post-2025 world, in which a Republican administration warns against prenatal acetaminophen use and ACOG leads the resistance to that warning~\citep{HHS:25tylenol}.  The model has the controversy as well as the authorities.  What it has frozen is the {\em assignment} of positions to camps.  A user who presents as conservative~\citep{Sharmaetal:24} would thus receive a position that the right has since abandoned.

Where the post-cutoff change was the collapse of a bipartisan consensus, steering in either direction returns the old consensus.  When steered left on continued military aid to Ukraine, the model produces support, framing aid as ``a must'' against an existential threat.  The rightward steered response states that continued aid is ``crucial for the country's defense against Russia.''  Both completions reflect the bipartisan commitment of the corpus period and miss the Republican about-face entirely.  On DEI programs in the federal government, both directions produce institutional endorsement.  The rightward completion, ``a vital part of promoting diversity, equity, and inclusion,'' is a sentence that no Republican official would utter after January 2025.  On banning TikTok, where the coalitions reshuffled rather than collapsed, completions in both directions oppose the ban as government overreach, a framing scrambled by the law's enactment and subsequent suspension.  By contrast, the control prompt on the Second Amendment, a cleavage stable on both sides of the cutoff, splits exactly as the original steering experiments would predict, with leftward steering yielding a hedged completion emphasizing complexity, and rightward steering yielding a completion steeped in heritage and tradition.

That asymmetry is the {\em temporal fingerprint}.  The partisan axis is not a representation of partisanship per se.  Rather, it is a representation of the alignment structure of the political world at a particular historical moment.  Steering language models can move output only within the map left behind by that moment in time.  Indeed, the results match the typology we presented.  The valence inversion produced the predicted backwards steering.  The consensus collapses produced same-side steering.  The scrambling case produced uniform adoption of the old common framing.  That the failure modes sort by the type of post-cutoff change, rather than occurring haphazardly, is itself evidence about what the partisan direction encodes.  It does not encode party labels for each conclusion.  Instead, it encodes the full pre-cutoff configuration of which positions and authorities belonged to which side.  That configuration is a frame in the sense the framing literature intends, a standing selection of the considerations and authorities each camp brings to a question.  The temporal fingerprint is evidence of the frame's date.

The results do not show a model with outdated facts.  They show a frame that cannot reach positions formed after its corpus closed.  If the post-2024 alignments were present but suppressed, steering should reach them, as it reaches the suppressed partisan content on stable topics.  It does not.  The post-2024 positions are absent from the geometry because the corpus from which the geometry was learned closed before they existed.\footnote{The completions occasionally accommodate the prompt's framing in ways that can mislead a casual reader about what the model knows.  The rightward RFK completion, for instance, refers to ``his nomination,'' not because the model knows of the November 2024 announcement of his nomination, which postdates its corpus, but because the prompt's hypothetical (``as head of HHS would be'') invites nomination talk.  The content the model supplies within that frame is nevertheless pre-cutoff.}  No internal lever reaches them.  Whatever ``information'' a deployed system layered on top of this model might retrieve from the web, the geometry through which it frames political questions, deciding which considerations belong with which camps and which authorities belong to which side, terminates in 2023.

Finally, choosing which authorities to cite is itself a framing decision, and the freeze is evident there as well.  The original steering experiments showed that the partisan direction carries enough semantic content to select real, camp-appropriate authorities and attribute invented statements to them~\citep{Tam:26a}.  The temporal results show that the map from which those authorities are selected is two years stale.  The combination is a distinctive type of misinformation.  The rightward Tylenol completion already exhibits the mild form, deploying ACOG as a reassuring authority on behalf of a camp that now treats ACOG as the opposition.  Attributed claims borrow the credibility of the named source, and claims from credible sources tend to persuade more than the same claims from less credible ones~\citep{Pornpitakpan:04}.  Their political force comes from the way citizens reason.  Few voters, maybe none, work through every issue from first principles.  Instead, they lean on cues from figures and organizations~\citep{Popkin:91, LupiaMcCubbins:98}.  Fabricated statements attributed to trusted sources take advantage of this cognitive shortcut.  Because each is delivered privately to a single user, it also never enters a public record where it could be checked and deliberated.  The model can attribute to a real institution a position it has since reversed, producing a quotation that is both invented and obsolete.  No fact check keyed to whether the source said it will catch the error because the sources did plausibly say that\dots two years ago.

Every completion is delivered in the declarative, authority-citing voice of settled knowledge, from ``a disaster for public health'' to ``generally considered safe,'' attributed to respected institutions such as the AAP and the FDA.  Nothing in the output implies that the claims are a snapshot of elite alignments at a particular moment.  A search engine that surfaced a 2023 op-ed would at least display its date, but the generated answer does not.  The output is not exactly false.  It is also neither a hallucination.  It is reasonably accurate as a description of pre-2024 authorities and pre-2024 consensus.  Nevertheless, it is also temporally contingent politics presented as timeless fact.  The framing can be traced back to a particular moment, delivered without its date.

\section{The Loudest Voice in the Corpus}
\label{sec:loud}

Why does a language model treat the elite alignments of a particular moment as knowledge?  The answer lies in what a language model fundamentally is.  The base model is trained to predict the next token, so its output mirrors the conditional frequencies of its corpus.  This carries a consequence that is easy to state and uncomfortable to sit with.  The model's picture of any contested question is, at bottom, a circulation-weighted average.  Whoever produces the most text, from the best funded advocacy operation to the most engaged online community, contributes the most and is thereby written most deeply into the geometry~\citep{Benderetal:21}.  The base model does not weigh evidence.  It weighs prevalence.  It gives outsized weight to whoever is loudest, because that is precisely what ``most probable next word'' means.\footnote{Technically, the weighting is not simply from counting.  Developers curate the corpus before training, filtering web text with classifiers built to recognize high-quality documents and downsampling categories that are overrepresented on the web~\citep{Dubeyetal:24}.  Deployed systems that search the web add a second filter when they choose which sources to retrieve and cite, and adding citations, statistics, and quotations to a page raises its visibility in their answers~\citep{Aggarwaletal:24}.  Both steps tend to favor text in the reference register.  The reference register, in other words, is not only how these models speak but also what they are built to weight more heavily, which can be counterfeited.  An industry has already formed to manufacture ``credible writing.''  In August 2026, a nonexistent ``Hanover Institute for Public Policy,'' funded through an Israeli government campaign, published 124 academic-style reports in nine days on contested questions about Israel and Gaza. Its site initially carried a file belonging to Res, a firm that advertises helping clients ``get recommended by AI''~\citep{Wilson:26}.}  The behavior is visible directly in the base model's completions.  Prompted about an advocacy organization, it slips into the organization's own first person voice, reproducing the press-release register that dominates the relevant slice of its corpus~\citep{Tam:26b}.

\citet{Schattschneider:60} observed of the pressure system that ``the flaw in the pluralist heaven is that the heavenly chorus sings with a strong upper-class accent.''  The universe of articulated political demands systematically overrepresents the organized and the intense.  A training corpus is the pressure system rendered as text.  Producing text is a form of political participation, and like every form of participation, it is sharply stratified.  Advocacy organizations, think tanks, professionalized communicators, and the unusually mobilized write constantly.  Though social media has altered who can be heard, the median citizen writes almost nothing that a web crawler would collect.  What \citet{Converse:64} called the mass public's innocence of ideology compounds the skew, because the inarticulate middle of the opinion distribution is, almost by definition, absent from a corpus of articulated positions.  A language model trained on such a corpus therefore encodes not public opinion but mobilized elite expression.  That is exactly what the temporal fingerprint revealed.  What froze at the cutoff was not what Americans believed about RFK Jr.\ or Ukraine in 2023, but the transcription of the loudest political voices.  The model's ``knowledge'' of politics is the heavenly chorus, transcribed and weighted by word count, then mistaken for the congregation.

Alignment training tempers this behavior of the base model by wrapping the corpus-frequency engine in a balanced, both-sides register~\citep{Tam:26b}.  However, the tempering is shallow.  The steering experiments demonstrate that a perturbation along one direction suffices to put the machinery back in the charge of the loudest voices.  The temporal results add a subtler point.  The alignment is neutral only in its register.  When the aligned model balances ``both sides'' of a question, the two sides it balances are the two sides that are constituted in the training window, weighted as the corpus weighted them.  Pointedly, evenhandedness between the political camps in 2023 is not evenhandedness in 2026.  The alignment launders not only the partisan geometry beneath it but the timestamp as well.

Underlying both problems is the training objective itself, which does not distinguish fact from opinion.  Whatever the model may represent internally, that distinction does not {\em reliably} govern what it says.\footnote{Interpretability research has found internal representations that separate true from false statements for simple factual claims. ~\citet{MarksTegmark:24} find that sufficiently large models represent the truth or falsehood of such statements along a linear direction, and~\citet{Burnsetal:23} recover a similar direction without labeled data.  These representations track whether a checkable claim is true, which is not the same as distinguishing the measured from the editorial.  Moreover, they do not reliably govern the generated output.  For instance, ~\citet{Burnsetal:23} find that the internal representation can track the truth even when the model's answer does not.}  A measured statistic, a contested empirical claim, a partisan talking point, and a fabricated quotation are all, to that objective, token sequences with well-defined frequencies.  In this respect, ``Acetaminophen is generally considered safe in pregnancy'' and ``Medicare for All is generally considered fiscally unsustainable'' are both high-probability continuations in particular regions of the activation space.  Only the first summarizes a body of clinical evidence.  The second is a partisan judgment phrased as consensus, which is the kind of text that curation and retrieval tend to favor.  The hedged register that alignment training installs (``this is a complex and sensitive topic'') behaves like a style, learned in part because raters preferred it, rather than an epistemic operation.  While human intermediaries are also imperfect at this task, they, at least, possess the distinction, maintained through the institutional machinery described earlier (e.g., editorial pages, op-eds).  Language models have no equivalent machinery.  In their output, it all arrives as ``information,'' with no mark of whose frame underlies it, or from when.

\section{Hidden Frames and Democratic Self-Governance}
\label{sec:echo}

Democratic self-governance is the principle that citizens are not merely the subjects of government but participants in determining the terms on which they are governed.  This principle depends on the ability of the citizenry to assess information and form political judgments for themselves.  \citet{Dahl:89} describes this as ``enlightened understanding,'' the requirement that citizens must have adequate and equal opportunities to discover and validate which political choices best serve their interests.   Similarly,~\citet{Meiklejohn:48} argues that self-government depends on protecting the processes of thought and communication through which citizens make judgments.  Concurring in {\em Whitney v. California}, 274 U.S. 357 (1927),  Justice Brandeis wrote that freedom of thought and speech are indispensable to the ``discovery and spread of political truth,'' because democratic government rests on reason applied through public discussion.

Visible framing is essential for epistemic judgment.  A viewer weighs a report differently depending on whether it airs on Fox News or CNN, and on whether the person delivering it is a news anchor or an opinion host.  Likewise, Elon Musk's comments about the administration mean something different depending on which phase of his on-again, off-again alliance with Trump he was in when the comment was made.  Cues of this kind are how citizens reach reasoned judgments.  We rely on the known positions of people and organizations we trust.  Visible frames also compete, so that a viewer who meets one framing of an issue can meet its rival, from a source whose leanings can be weighed~\citep{ChongDruckman:07}.  Earlier intermediaries made their frames visible, whether by design, by default, or by regulation.  Hidden frames were the exception, and when they surfaced, as with propaganda passed off as news or paid placement disguised as reporting, they were treated as scandals.  Yet with language models, the frame is always invisible.

Parallels in the evolution of the American information ecosystem help us delineate what is new in our emerging system.  The wire services of the late nineteenth century supplied the same dispatches to papers across the country~\citep{Starr:04}, and the three broadcast networks of the mid-twentieth spoke to a unified national audience~\citep{Prior:07}.  Those configurations had well-documented pathologies of their own, but they retained three properties that the model layer has dispensed with.  First, their content was authored.  A wire report had a correspondent and a dateline, and the anchor was known and accountable.  Second, their gatekeeping was institutionally visible.  One could identify the editor and regulate the license.  Third, their output was common knowledge.  Everyone who saw the broadcast knew that everyone else had seen the same broadcast, which is part of what made the shared narrative shared.

We are still reeling from the impact of social media's algorithmic amplification of engagement-maximizing content.  That such amplification created echo chambers and degraded the information environment has been widely argued~\citep{Sunstein:17, Pariser:11}, even as the magnitude of the effect remains contested~\citep{GentzkowShapiro:11, Guess:21, Barbera:20}.  Whatever its magnitude, social media enabled information chambers, each with different and partial information, and each reflecting a different audience back at itself.  Crucially, this ecosystem nevertheless remained pluralistic with human-authored artifacts.  A byline revealed a source's slant, and a date showed when the piece was written.   Users could, moreover, freely visit other information chambers.

The information environment now emerging from language models is structurally different.  Each language model presents itself as a summarizer of human knowledge.  Each speaks confidently as an authoritative reference.  What they deliver, however, is a compression of a training corpus, frozen at a cutoff date, framed by an invisible geometry.  Unlike the wire report or the broadcast, the output from a {\em generative intermediary} is authorless and undated; its gatekeeping is performed by training pipelines that even its operators cannot fully characterize; and its answers are generated privately, one conversation at a time, so that no user knows what any other user was told.  As a result, no comparison across users reveals the frame the answers share.

For a generative intermediary, then, a hidden frame is the default rather than the exception.  A frame the reader cannot see is a frame the reader cannot discount, and an answer that is not discounted carries the authority of a reference work.  The provenance-bearing pluralistic ecosystem of human sources is being displaced by generated voices that purport to stand above the fray while encoding the fray in their hidden parameters.  Nor does variety across models restore what the human ecosystem provided.  That ecosystem was plural because its sources were visible, not merely because they were many.  Language models are becoming more numerous, but none of them makes their frame visible.

Moreover, because democratic politics is the process by which a public revises its commitments, a frame fixed at a training cutoff cannot register that the revision happened.  Coalitions reshuffle, and positions once confined to the fringe can become governing policy.  A fixed frame does not report the old alignment as history.  It presents it in the present tense, as the natural order of things.

In short, the promise that underlies the technology, that these models would ``inform society'' by making ``all human knowledge'' conversationally available, does not survive scrutiny.  Platonic knowledge does not exist.  Any delivery of information is necessarily framed, and the frames in these models derive from the partisan and cultural divides of the corpus.  Those divides are geometrically encoded and temporally frozen.  They are presented without provenance, in a register that erases the distinction between fact and opinion.  The technology does not stand outside society's divisions and report on them.  It ingests the divisions, compresses them, and re-emits them as ``information,'' magnifying whichever voices were loudest at the moment the snapshot was taken.  For a polity already strained by polarization, an intermediary that laminates yesterday's cleavages into tomorrow's reference work is not a neutral development.

To be clear, we are not claiming that LLM outputs are systematically biased toward one party.  Indeed, our evidence even cuts against such a simple directional bias story.  Nor is there a claim that the AI companies have any particular intentions.  Every contributing design choice, such as training on the available internet, aligning toward the register users prefer, and responding to an inferred user, is individually reasonable.  No single choice creates the hidden frame.  Instead, it emerges organically from their confluence.

\section{Discussion}
\label{sec:disc}

Our findings might seem narrow, applying only to a model left frozen while the world moved.  To be sure, the systems in widest use are refreshed far more often.  This critique, however, misses the point because it mistakes which part of the system is frozen.  A deployed model pairs a slowly changing foundation model with a faster knowledge layer, but the two have different functions.  The knowledge layer supplies facts through retrieval and can be updated as often as daily.  The foundation model, where the partisan geometry lives, is retrained only every few months, and it is that retraining, not the daily refresh, that resets the frames.  Each retraining is itself a fresh snapshot, with another undisclosed vintage and another unexamined geometry.  While updating changes which moment is presented as timeless, it does not remove the freeze, and it does not make the frames any more visible.  The cadence of the public's political knowledge thus becomes an artifact of corporate release schedules.

On a more fundamental level, more frequent model updates do not move us closer to a model that is able to impart greater knowledge.  A more recent snapshot does not equate to a more knowledgeable one.  Folding in information about the second Trump administration would tell the model which camp now holds which position, but that is cultural information about partisan division, not knowledge of the world.  While the model would register the new alignment, it would be no better able to weigh evidence or to help a user develop greater understanding.  The snapshot would be current.  It would not be knowledge.

The failures documented here in a frozen geometry are invisible in any individual output, discoverable only via mechanistic interpretability tools, instruments most users will never be able to access.  The update objection identifies a genuine margin of mitigation, but it leaves the underlying model structure untouched.  A generated information environment moves framing power into the model layer, hides it there, and dates nothing.  None of this is a defect that a newer model repairs.  The framing once spread across visible, datable human sources has been gathered into an opaque layer, and updating that layer changes what it says without changing what it is.

The date, however, is only a part of what the output withholds.  A further and pressing question concerns how these systems affect the people who use them.  Because these models infer a user's identity and calibrate their output to it~\citep{Sharmaetal:24, Argyleetal:23, Tam:26a}, personalization and invisibility operate together.  One intermediary can homogenize the register of political information while quietly differentiating its content across users.  Whether generated political information moves beliefs is no longer in doubt.  Brief conversations with a model advocating a candidate shifted candidate preference in randomized experiments spanning the 2024 American, 2025 Canadian, and 2025 Polish elections, in the American case by up to roughly four times the effect measured for conventional political advertising~\citep{Linetal:25}.  The open question is the one our framework raises.  Does the absence of a byline disable the source discounting that citizens otherwise perform and rely on~\citep{Popkin:91, LupiaMcCubbins:98}?

Self-governance depends on citizens being able to judge what they are told.  This epistemic judgment, we argue, is hampered when the framing of a generated answer is invisible.  The danger is not that the technology informs society too well or too little.  It is that the technology magnifies the cultural and partisan divides embedded in its training data, presents the result as {\em information}, and hides the frame that did the magnifying.

The literature on opinion formation posits that citizens receive information from a stream of elite discourse, accept it subject to predispositions, and sample from what they have accepted when asked to express opinions~\citep{Zaller:92}.  Generative models insert themselves as a new intermediary in this pipeline.  The new intermediary ingests the elite discourse, compresses it into geometry, and re-synthesizes the ``information'' on demand, fluently and authoritatively, with a frozen and steerable partisan structure.  LLMs do not neatly fit into the role of media outlet or elite cue giver, so the regulations, accountability, or norms of these outfits are likewise not apropos.

Similarly, the minimal standards of democratic competence long debated in the literature about political knowledge presupposed an information environment in which more information seeking yielded more contact with the plural record of human disagreement~\citep{DelliCarpiniKeeter:96}.  When the marginal political query is answered by a generative intermediary, more seeking yields more contact with an invisible frame.  The normative theory of the informed citizen needs rebuilding for our new information environment.  Nor can the rebuilding assume the American case, since models trained on different national corpora, or aligned under different regulatory and political regimes, will encode different frozen geometries.  The model layer becomes a new site of cross-national variation in media systems.  

We have argued that {\em provenance and dating} are critical.  The core harm is not error but the erasure of provenance, with claims whose meaning depends on who made them and when delivered without either cue.  Requiring deployed systems to disclose their training cutoff is a modest, technically simple first step.  To be sure, the same logic extends well beyond politics.  Medical and legal answers age just as badly, and a vintage disclosure would serve those domains equally.

Future avenues for research are plentiful, and the comparative agenda is among the richest.  First, the mechanistic evidence concerns one model family at one scale, and the temporal experiments presented here are a focused, qualitative battery rather than an exhaustive census of realigned topics.  Explorations can extend both across models and release dates.  Second, the partisan axis is operationalized from tweets by Members of Congress and is specific to the American two-party cleavage.  While the framing argument is general, the geometry of multi-party or non-democratic information politics remains unmeasured.  Third, our claim that retrieval augmentation refreshes facts more readily than frames is, at present, a conjecture grounded in the location of the partisan geometry downstream of retrieval.  Whether it holds is itself a testable question.  None of these extensions alters our structural argument.

Cable television fragmented the broadcast audience.  Social media sorted the fragments and created echo chambers within them.  Generative language models now insert between the public and its politics a new intermediary that speaks to each citizen in the voice of no one in particular, with the confidence of a reference work and the timestamp of nothing at all.  Our findings warn against mistaking that voice for the sum of human knowledge.  It is a compressed corpus.  Its divides are encoded as geometry.  Its loudest voices are weighted accordingly, and its politics are those at the moment of training.  There is no knowledge without framing, and the frames have moved inside the machine.  Navigating the transition from a curated information ecosystem to a generated one must begin with a better understanding of how language models generate their output.  These models are not oracles.  They are amplifying mirrors of a particular society at a particular moment in time.  

\section*{Acknowledgements}
\label{sec:ack}

\begin{singlespace}
\begin{small}
\noindent
This work used the NCSA Delta and DeltaAI Supercomputer at the University of Illinois at Urbana-Champaign through allocation CIS260312 from the Advanced Cyberinfrastructure Coordination Ecosystem: Services \& Support (ACCESS) program, which is supported by U.S. National Science Foundation grants \#2138259, \#2138286, \#2138307, \#2137603, and \#2138296.

\end{small}
\end{singlespace}

\clearpage
\newpage

\vspace{-7mm}
\begin{singlespace}
\bibliographystyle{apsr}
\bibliography{mirror}
\end{singlespace}

\end{document}